\documentclass[aps,prx,twocolumn,superscriptaddress,centertags,floatfix,longbibliography]{revtex4-2}

\usepackage{mathptmx}

\usepackage{amsmath}
\usepackage{amssymb}
\usepackage{amsthm}

\newtheorem{proposition}{Proposition}
\theoremstyle{definition}

\theoremstyle{remark}

\usepackage{booktabs}
\usepackage{graphicx}
\usepackage{xcolor}

\usepackage{algpseudocode}
\makeatletter
\newcounter{algorithm}
\def\thealgorithm{\arabic{algorithm}}
\def\fps@algorithm{htbp}
\def\ftype@algorithm{16}
\def\ext@algorithm{\jobname.loa}
\def\fnum@algorithm{Algorithm \thealgorithm}
\newenvironment{algorithm}
  {\@float{algorithm}}
  {\end@float}
\@namedef{algorithm*}{\@dblfloat{algorithm}}
\@namedef{endalgorithm*}{\end@dblfloat}
\makeatother

\usepackage[colorlinks,linkcolor=blue,citecolor=blue,urlcolor=blue]{hyperref}

\setcitestyle{super,comma}

\begin{document}

\title{Causal discovery from Bayesian symbolic regression}

\author{Roger Guimer\`a}
\email{roger.guimera@urv.cat}
\affiliation{Department of Chemical Engineering, Universitat Rovira i Virgili, Tarragona, Spain}
\affiliation{Center for Computational Science and Applied Mathematics (ComSCIAM), Universitat Rovira i Virgili, Tarragona, Spain}
\affiliation{ICREA, Barcelona, Spain}

\author{Kai-Hendrik Cohrs}
\affiliation{Image Processing Laboratory (IPL), Universitat de Val\`encia, Val\`encia, Spain}

\author{Sergio Pedreira}
\affiliation{Department of Chemical Engineering, Universitat Rovira i Virgili, Tarragona, Spain}

\author{Marta Sales-Pardo}
\email{marta.sales@urv.cat}
\affiliation{Department of Chemical Engineering, Universitat Rovira i Virgili, Tarragona, Spain}
\affiliation{Center for Computational Science and Applied Mathematics (ComSCIAM), Universitat Rovira i Virgili, Tarragona, Spain}

\author{Gustau Camps-Valls}
\email{gustau.camps@uv.es}
\affiliation{Image Processing Laboratory (IPL), Universitat de Val\`encia, Val\`encia, Spain}

\begin{abstract}
Causal discovery from observational data is central to scientific reasoning, yet established constraint-based and score-based methods often recover only a Markov equivalence class and provide no signal in bivariate or strongly autocorrelated regimes typical in some areas of science. Here, we recast causal discovery as a Bayesian symbolic regression problem. Given a set of candidate parents for each target variable, the Bayesian symbolic regression samples closed-form expressions from a posterior over symbolic models. The fraction of sampled expressions in which each candidate variable appears defines a posterior inclusion probability that we use as the causal score. This replaces pointwise model selection by an ensemble of plausible mechanisms. We evaluate the method on coupled nonlinear bivariate dynamics, on the CauseMe benchmark suite spanning linear, chaotic, nonlinear, higher-dimensional and high-noise settings, and on a four-variable biogeoscience structural causal model under Gaussian, heavy-tailed and mixed-noise regimes. Our approach is competitive with, and often outperforms, specialized causal discovery algorithms. Additionally, the method returns closed-form mechanisms leading to explicit causality and interpretability.
\end{abstract}

\maketitle

%\noindent\textbf{Keywords:} causal discovery, symbolic regression, Bayesian machine scientist, minimum description length, identifiability, Earth and climate science.\\

\section{Introduction}
\label{sec:intro}
Scientific understanding is built on \emph{mechanisms} that describe how interventions on one part of a system propagate through the rest. Moving from statistical association to structural causation is a long-standing endeavor that has reshaped many disciplines~\cite{Pearl09,peters2017elements,Pearl2000}. In disciplines where true experiments are seldom possible, causal reasoning is essential to attribute change to forcings, to identify indirect connections, and to evaluate process-based and data-driven models on physical and mechanistic rather than purely predictive grounds. This is the case for Earth and climate sciences~\cite{Runge19natcomm,Reichstein19nat,CampsValls23physcausaldiscovery}, neuroscience~\cite{reid19,siddiqi2022causal}, epidemiology~\cite{hernan2020causal}, and economics~\cite{hicks1980causality}, to name a few examples.

Dominant causal discovery paradigms rely on conditional-independence tests over a presumed faithful distribution and return graphs only up to a Markov equivalence class~\cite{Spirtes2000,peters2017elements}. In the bivariate setting, where one wishes to distinguish $X\to Y$ from $Y\to X$, the marginal independence structure is the same for both directions and constraint-based methods provide no signal. To break this symmetry, one can exploit structural restrictions on the functional form (additive noise models~\cite{Hoyer09,mooij2016distinguishing}, LiNGAM~\cite{shimizu2006linear}), invariance under environments~\cite{peters2016causal,heinze2018invariant}, or algorithmic information principles~\cite{janzing2012information,sun2008causal,reichenbach1991direction}. Approaches in the latter family can assign complexity to causal graphs---usually measured as the causal graph Kolmogorov complexity approximated through kernel embeddings~\cite{Mitrovic18kcdc,PerezSuay19geoscience} or stochastic codes~\cite{budhathoki2017causal,Marx19slope}---and can therefore use the minimum description length (MDL) principle for causal graph selection. Still other approaches operate in the time domain through Granger causality~\cite{Granger69}, convergent cross mapping~\cite{Sugihara12}, and conditional-independence variants tailored to autocorrelation~\cite{runge2019detecting,runge2020causality}.

Here, we argue that probabilistic symbolic regression approaches \cite{guimera20,guimera26} offer a better solution to causal discovery, even in the bivariate case. Indeed, from dynamical systems, we know that the equilibrium states of a first-order ordinary differential (ODE) system on which we intervene can be recovered using a structural causal model (SCM)~\cite{mooij2013ordinary}. Interestingly, under standard stability assumptions, recovering the differential (or difference) equation that governs a process therefore amounts to recovering the SCM of its equilibrium or stationary regime. This result suggests that symbolic regression is then not only a predictive surrogate but a {\em constructive} route to the structural mechanisms targeted by causal inference~\cite{CampsValls23physcausaldiscovery}. 

With this motivation in mind, here we use methods that explore the space of closed-form mathematical expressions to find causal relationships. In particular, we use the Bayesian machine scientist (BMS)~\cite{guimera20}, which explores this space with a Markov chain Monte Carlo (MCMC) sampler over symbolic expressions and assigns each expression a plausibility based on its description length~\cite{bmscode}. The output is a sample taken from the posterior distribution of plausible mechanisms. We define the causal score of a candidate parent variable in the causal graph as the variable's posterior \emph{inclusion frequency}, that is, the fraction of sampled expressions in which the variable appears. This score handles many candidate parents and lagged predictors, is robust to high noise levels, accommodates autocorrelation, and remains informative in regimes where several expressions fit the data. Besides, our method simultaneously delivers a causal graph and explicit functional forms for plausible mechanisms. 

This paper makes three contributions. First, a formalization of causal discovery as Bayesian symbolic regression, in which the causal score of each candidate parent is its posterior inclusion frequency in the ensemble of sampled mechanisms. Second, an identifiability analysis that connects causal recovery to the recently characterized learnability transition of probabilistic model selection~\cite{fajardo2023fundamental}, separating a learnable low-noise regime, a transitional regime, and an unlearnable high-noise regime. Third, an evaluation on a wide variety of problems, from bivariate nonlinear dynamics, to biogeoscience structural causal models, and standard benchmark suites~\cite{runge2020causality,runge2019detecting}. In all of them causalBMS attains the highest or very competitive performance compared to specialized methods elsewhere, while additionally returning the closed-form mechanism that produced the data or very good approximations to those. 

\section{Causal scores from Bayesian symbolic regression}
\label{sec:method}

\subsection{Bayesian symbolic regression from time series}
\label{sec:bms}

We consider a vector of observable variables $\mathbf{x}(t)\in\mathbb{R}^{d}$, $t=1,\dots,T$. Each component $x^{i}$ is assumed to be generated by a structural equation\footnote{For simplicity, here we ignore possible contemporaneous effects, but we consider a case where such effects exist later in our experimental results section.}
\begin{equation}
x^{i}(t) \;=\; f^{*}_{i}\!\left(\mathbf{x}_{\mathrm{Pa}(i)}(t-1),\dots,\mathbf{x}_{\mathrm{Pa}(i)}(t-\tau_{\max})\right) \;+\; \varepsilon^{i}(t),
\label{eq:scm}
\end{equation}
with $\mathrm{Pa}(i)\subseteq\{1,\dots,d\}$ the set of causal parents of $x^{i}$ and $\varepsilon^{i} \sim \mathcal{N}(0, \sigma)$ an exogenous Gaussian noise term assumed mutually independent across $i$ and $t$. In this setting, the goal of causal discovery is to recover $\{\mathrm{Pa}(i)\}_{i=1}^{d}$, or its lagged extension, from the data $\mathcal{D}=\{\mathbf{x}(t)\}_{t=1}^{T}$. In contrast, the goal of symbolic regression is originally to recover $f^{*}_{i}$ in closed form. The key observation is that, under the stated conditions, causal discovery becomes easier if one first solves the symbolic regression problem.

Indeed, we note that Eq.~\eqref{eq:scm} also admits a structural causal interpretation. For a deterministic dynamical system $\dot{\mathbf{x}}=\mathbf{F}(\mathbf{x})$ with a globally stable equilibrium, the parent set induced by $\mathbf{F}$ at equilibrium defines an SCM that is consistent with the do-calculus on perfect interventions~\cite{mooij2013ordinary}. The same construction applies to discrete-time systems satisfying Eq.~\eqref{eq:scm} under standard stationarity conditions. Recovering $f^{*}_{i}$ therefore yields, under these assumptions, both a predictive equation and the structural mechanism that defines the causal effect of intervening on the parents of $x^{i}$. This intuition that symbolic regression drives causal discovery can thus be properly formalized. 

The space of admissible expressions for the $i$-th component is denoted $\mathcal{F}_{i}$. Following standard practice in symbolic regression~\cite{guimera20,fajardo2023fundamental,guimera26},  we represent closed-form expressions $f_i\in\mathcal{F}_{i}$ as labeled trees whose internal nodes are operators drawn from a fixed alphabet $\mathcal{O}=\{+,\times,\exp,\log,\sin,\dots\}$ and whose leaves are variables or real parameters $\boldsymbol{\theta}\in\Theta_{f_i}\subseteq\mathbb{R}^{k_{f_i}}$. Such alphabet is often selected (restricted) by domain knowledge and process understanding. In symbolic regression approaches such as SINDy~\cite{brunton2016discovering}, the model is restricted to linear combinations of certain basis functions. The selection of these basis functions is critical and, in general, limits the expressiveness of the approach~\cite{cabanastirapu25}. 
By contrast, probabilistic symbolic regression and the BMS~\cite{guimera20,guimera26} impose no such restrictions and allow any function combination and composition from the alphabet.

For a target $x^{i}$ and an expression $f_i\in\mathcal{F}_{i}$, the BMS uses probabilistic arguments to assign to each candidate expression $f_i\in\mathcal{F}_{i}$ the posterior probability
\begin{eqnarray}
p(f_i{}|{}\mathcal{D}) & = & \frac{1}{p(\mathcal{D})}\int_{\Theta_{f_i}} p(\mathcal{D}{}|{}f_i,\boldsymbol{\theta})\,p(\boldsymbol{\theta}{}|{}f_i)\,p(f_i)\,\mathrm{d}\boldsymbol{\theta} \nonumber\\
& = &  \frac{\exp\!\big[-\mathcal{H}(f_i)\big]}{Z},
\label{eq:posterior}
\end{eqnarray}
with description length (or information Hamiltonian \cite{enslin19}) $\mathcal{H}(f_i)=-\log p(\mathcal{D},f_i)$, evidence (or partition function) $Z=p(\mathcal{D})$, parameter prior $p(\boldsymbol{\theta}{}|{}f_i)$, and structural prior $p(f_i)$~\cite{guimera20,grunwald07}. The structural prior can either be learned from a corpus of mathematical expressions extracted from scientific texts under a maximum-entropy assumption~\cite{guimera20} or chosen to encode other forms of knowledge about the target expression \cite{fox2024}. Under the assumption of Gaussian exogenous noise $\varepsilon^{i}(t)\sim \mathcal{N}(0, \sigma)$ and using Laplace's approximation around the maximum-likelihood estimator $\boldsymbol{\theta}^{*} = \arg \max_{\boldsymbol{\theta}} p(\mathcal{D}{}|{}f_i,\boldsymbol{\theta_{f_i}})$, the description length is
\begin{equation}
\mathcal{H}(f_i) \;\approx\; \frac{B(f_i)}{2} \;-\; \log p(f_i),
\label{eq:bic}
\end{equation}
where $B(f_i)=-2\log p(\mathcal{D}{}|{}f_i, \boldsymbol{\theta}^{*})+(k_{f_i} + 1)\log T$ is the Bayesian information criterion~\cite{schwarz78} of $f_i$, $k_{f_i}$ is the number of parameters of $f_i$ \footnote{The +1 term corresponds to one additional parameter of the probabilistic model, namely, the standard deviation $\sigma$ of the exogenous noise.}, and $T$ is the number of observed time points. Note that Eq.~\eqref{eq:bic} makes the connection with the MDL two-part code\cite{grunwald07} explicit: $B(f_i)/2$ encodes data and parameters, while $-\log p(f_i)$ encodes the structural complexity of the expression.

The BMS explores the space $\mathcal{F}_{i}$ by generating a Metropolis chain over expression trees with three local moves: root addition or removal, elementary-tree replacement, and node replacement~\cite{guimera20}. Parallel tempering across a temperature ladder $\{T_{k}\}$ stabilizes the exploration of the rugged $\mathcal{H}$-landscape~\cite{guimera20}, and the chain returns a set $\mathcal{S}_{i}=\{f^{(j)}_{i}\}_{j=1}^{M}$ of expressions sampled from $p(f_i{}|{}\mathcal{D})$. 
Note that this is a key aspect of our approach, as averages over the chain of models $\mathcal{S}_{i}$ are an approximation to an average over the posterior, which is in general a better estimator of any observable than those provided by any single model \cite{guimera26}.

\subsection{From sampled expressions to causal scores}
\label{sec:scores}

Each sampled expression $f^{(j)}_{i}\in\mathcal{S}_{i}$ specifies a set of parents that plausibly enter the causal mechanism for $x^{i}$. Pooling the expressions in the chain $\mathcal{S}_{i}$ yields a posterior estimate of the probability that a candidate parent participates in the generating equation.
Specifically, for a target $x^{i}$ and a candidate parent $x^{k}(t-\tau)$, the posterior inclusion frequency is
\begin{equation}
{\pi}_{k\to i}(\tau) \;=\; \frac{1}{M}\sum_{j=1}^{M}\mathbb{I}\!\left[x^{k}(t-\tau)\in \mathcal{L}\big(f^{(j)}_{i}\big)\right],
\label{eq:inclusion}
\end{equation}
where $\mathcal{L}(f)$ denotes the set of leaves of expression $f$ that are observable variables, $\mathbb{I}$ is the indicator function, and the sum is over the sampled expressions in $\mathcal{S}_{i}$. 
Marginalizing over lags, we obtain a variable-level score independent of lag
\begin{equation}
{\pi}_{k\to i} \;=\; \frac{1}{M}\sum_{j=1}^{M}\mathbb{I}\!\left[\exists\,\tau:\, x^{k}(t-\tau)\in\mathcal{L}\big(f^{(j)}_{i}\big)\right],
\label{eq:inclusion_marg}
\end{equation}
which we use as the score of the inferred causal relationship $k\to i$. By thresholding ${\pi}_{k\to i}$ for all variables, we obtain a directed causal graph; by ranking edges in the causal graph by ${\pi}_{k\to i}$, we obtain precision-recall and ROC curves. The full procedure is summarized in Algorithm~\ref{alg:causalbms}.
\begin{algorithm}[t!]
\caption{causalBMS: Causal discovery from Bayesian symbolic regression}
\label{alg:causalbms}
\begin{algorithmic}[1]

\Require Data $\mathcal{D}=\{\mathbf{x}(t)\}_{t=1}^{T}$; alphabet $\mathcal{O}$; corpus-derived prior $p(f)$; maximum lag $\tau_{\max}$; chain length $M$; tempering ladder $\{T_{k}\}_{k=1}^{K}$.
\Ensure  Causal score matrix $\hat{\Pi}\in[0,1]^{d\times d}$ and ensembles of equations $\{\mathcal{S}_{i}\}_{i=1}^{d}$.
\For{$i=1$ \textbf{to} $d$}
   \State Construct candidate predictor set $\mathcal{X}_{i}=\{x^{k}(t-\tau):\,k\!\in\!\{1,\dots,d\},\,\tau\!\in\!\{0,\dots,\tau_{\max}\}\}\setminus\{x^{i}(t)\}$.
   \State Run parallel-tempered Metropolis MCMC on $\mathcal{F}_{i}$ targeting $p(f{}|{}\mathcal{D})\propto \exp[-\mathcal{H}(f)]$ via Eq.~\eqref{eq:bic}.
   \State Collect post-burn-in samples $\mathcal{S}_{i}=\{f^{(j)}_{i}\}_{j=1}^{M}$ from the cold replica.
   \For{$k=1$ \textbf{to} $d$, $k\ne i$}
      \State ${\pi}_{k\to i} \gets \tfrac{1}{M}\sum_{j}\mathbb{I}\!\left[\exists\tau:\,x^{k}(t-\tau)\in\mathrm{Var}(f^{(j)}_{i})\right]$.
   \EndFor
\EndFor
\State \Return ${\Pi}=({\pi}_{k\to i})_{k,i}$ and the ensembles $\mathcal{S}_{i}$.
\end{algorithmic}
\end{algorithm}

\subsection{Consistency, equation learnability and identifiability of causal parents}
\label{sec:identifiability}

The description length is a consistent model-selection criterion, meaning that, as $T\to\infty$, the posterior $p(f_i{}|{}\mathcal{D})$ concentrates on the model that minimizes $\mathcal{H}(f_i)$ over $\mathcal{F}_{i}$, which happens to be the ground-truth model $f^{*}_{i}$ as long as $f^{*}_{i} \in \mathcal{F}_{i}$  \cite{fajardo2023fundamental,grunwald07,schwarz78}. We summarize the implications for the inclusion-frequency score.

\begin{proposition}[Asymptotic recovery of the causal parents]
\label{prop:consistency}
Assume that (i) the true mechanism $f^{*}_{i}$ is in $\mathcal{F}_{i}$; (ii) the prior $p(f_i)$ assigns strictly positive mass to $f^{*}_{i}$; (iii) the exogenous noise $\sigma$ is finite; and (iv) the MCMC sampler is ergodic. Then ${\pi}_{k\to i}\to \mathbb{I}\!\left[k\in\mathrm{Pa}(i)\right]$ with probability approaching one as $T\to\infty$ and $M\to\infty$.
\end{proposition}

For a fixed and finite $T$, however, recovery is not guaranteed at all noise levels even when assumptions {\em (i)}, {\em (ii)}, and {\em (iv)} above hold. Indeed, Fajardo-Fontiveros et al.~\cite{fajardo2023fundamental} show that probabilistic model selection over closed-form expressions exhibits a sharp transition between a low-noise regime in which the true model has the smallest description length and a high-noise regime in which a trivial model becomes more plausible. Since probabilistic model selection is Bayes optimal, no other method will be able to recover the true generating model in the high-noise regime. An upper bound (that is typically tight) to the transition noise is
\begin{equation}
\sigma^{\times} \;\approx\; \left(\frac{\langle\delta^{2}\rangle\, T}{2\Delta^{*}_{\!M}+(k_i^{*}-1)\log T}\right)^{1/2},
\label{eq:phase_bound}
\end{equation}
where $k_i^{*}$ is the number of parameters of $f_i^{*}$, $\Delta^{*}_{\!M}=\log p(f_i^{c}) - \log p(f_i^{*})$ is the gap in structural description length between $f_i^{*}$ and the trivial model $f_i^{c}=\arg\max_{f_i} p(f_i)$ (which, in our maximum entropy prior corresponds to a constant $f_i^c = \theta_0$), and $\langle\delta^{2}\rangle$ is the variance of $f_i^{*}$ over the support of the predictors~\cite{fajardo2023fundamental}. This leads us to a second proposition.

\begin{proposition}[Assymptotic causal unidentifiability]
\label{prop:unlearnable}
For finite $T$ and $\sigma \to \infty$, the posterior $p(f_i{}|{}\mathcal{D})$ concentrates around the trivial model $f_i^c=\arg \max_{f_i} p(f_i)$, the inclusion frequencies ${\pi}_{k\to i}$ approach the prior expectations $\mathbb{E}_{p(f_i^c)}\big[\mathbb{I}[x^{k}\in\mathcal{L}(f_i^c)]\big]$ independently of the true $\mathrm{Pa}(i)$, and the causal score loses all discriminative power.
\end{proposition}

Yet, whether causal parents can be detected at finite $\sigma > \sigma^{\times}$, where the exact true model cannot possibly be identified from the data alone, is an open empirical question---in principle, it is possible that, even when the ground-truth model $f_i^*$ is unlearnable, some other model with the correct causal structure maximizes the posterior. We return to this issue below in section \ref{sec:bivariate}.

% =======================================================================
\section{Experimental results}
\label{sec:results}

\begin{figure*}[t!]
\centering
\includegraphics[width=\textwidth]{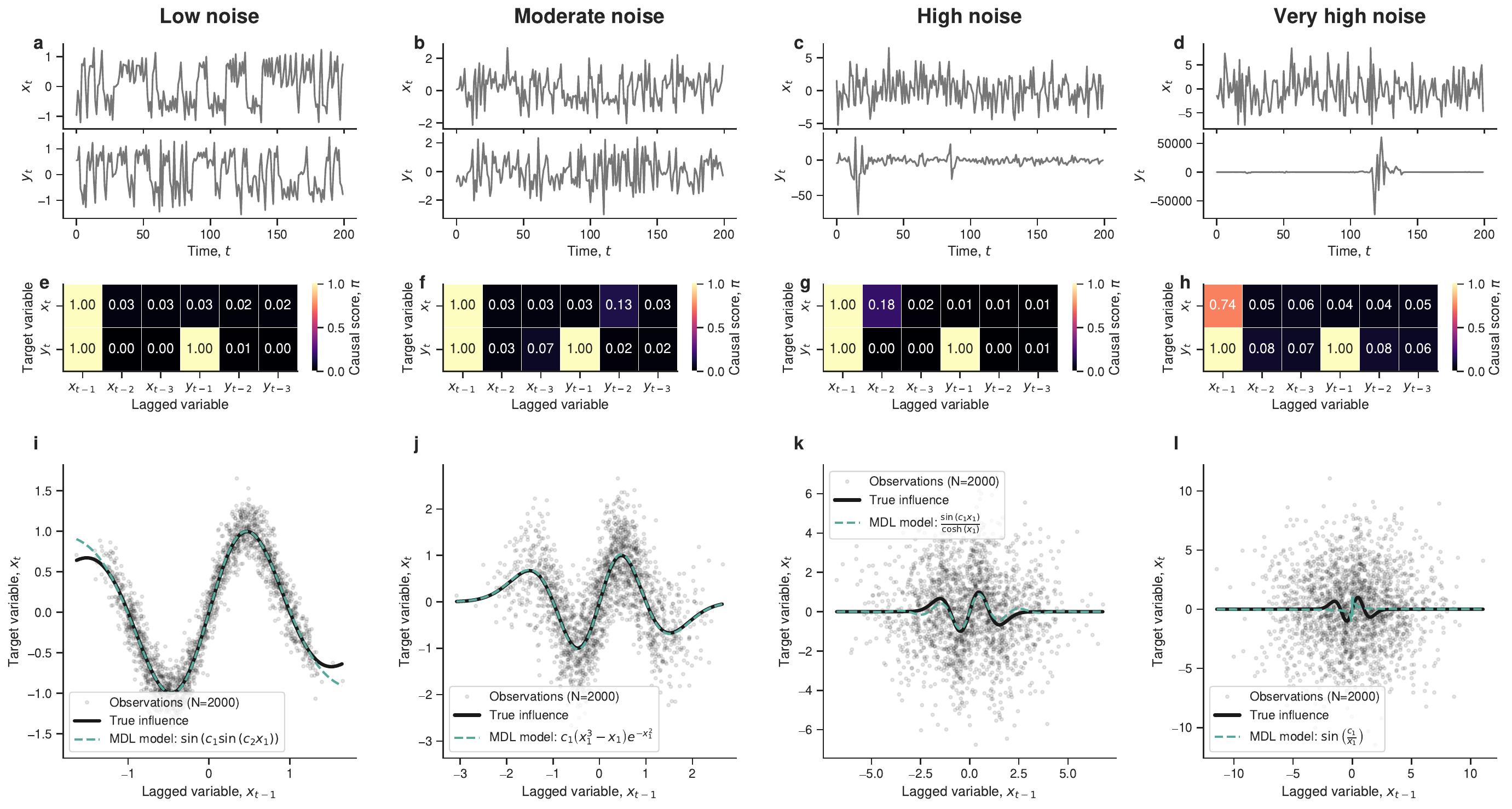}
\caption{\textbf{Causal and functional recovery in unidirectionally coupled nonlinear dynamics.} 
(a-d) Sample observed trajectories $x_t$ and $y_t$ for the system in Eq.~\eqref{eq:coupled}.
(e-h) Causal scores ${\pi}_{k\to i}(\tau)$ obtained by causalBMS from Eq.~\eqref{eq:inclusion}.
(i-l) Minimum description length expressions sampled by causalBMS (green) reproduce the true self-influence (black). Grey points are the $N=2000$ observations. Equation labels in the legend are the closed-form expressions inferred by the chain.
We consider four noise regimes: low noise ($\sigma=0.04$, first column); moderate noise ($\sigma=0.4$, second column); high noise ($\sigma=4$, third column); very high noise ($\sigma=10$, fourth column).
}
\label{fig:shape}
\end{figure*}
We evaluate our causal discovery approach based on BMS (causalBMS) across a wide range of datasets, to understand how the approach performs under relevant scenarios of nonlinear processes, strong couplings, non-stationarity, high-dimensional scenarios, and different levels and diversity of noise sources. All aspects have been identified as critical for causal discovery \cite{Runge19natcomm}. In particular, we evaluate the method on: (i) a coupled nonlinear bivariate dynamical system; (ii) the CauseMe benchmark suite spanning linear, chaotic, nonlinear, high-dimensional and high-noise settings; and (iii) a biogeoscience structural causal model of shortwave radiation, air temperature, gross primary productivity and ecosystem respiration under Gaussian, heavy-tailed and mixed-noise regimes. 

We compare causalBMS to standard methods in the literature (PCMCI$^{+}$~\cite{runge2019detecting,krich2020estimating}, VAR-LiNGAM~\cite{shimizu2006linear,hyvarinen2010estimation}, convergent cross mapping (CCM)~\cite{Sugihara12}, and a wide variety of methods in the \href{https://causeme.uv.es}{CauseMe} platform leaderboard~\cite{runge2020causality}), and through standard metrics of performance (F1 score, precision-recall scores and the ROC AUC metric). We also study the robustness of symbolic regression to noise in the coupled nonlinear bivariate dynamical system and in the biogeoscience selected case study.

The experimental setup requires running causalBMS multiple times and averaging the results. Computational cost is dominated by the MCMC sweep across targets; in our experiments we use a minimum of $M\!\sim\!10^{4}$ retained samples after burn-in, parallel tempering with a minimum of $K=10$ replicas, and a tree-size cap $|f_i|\le 50$ nodes. 

% ---------------------------------------------------------------------------
\subsection{Causal and functional recovery in unidirectionally coupled nonlinear dynamics}
\label{sec:bivariate}
% ---------------------------------------------------------------------------

We start by considering a bivariate coupled system that combines self-excitation with a unidirectional coupling and is governed by the following equations
\begin{align}
x_{t} & = 3.4\,x_{t-1}(1-x_{t-1}^{2})\exp(-x_{t-1}^{2}) + \varepsilon^{x}_{t}, \nonumber\\
y_{t} & = 3.4\,y_{t-1}(1-y_{t-1}^{2})\exp(-y_{t-1}^{2}) + 0.5\,x_{t-1}\,y_{t-1} + \varepsilon^{y}_{t},
\label{eq:coupled}
\end{align}
with $\varepsilon\sim\mathcal{N}(0, \sigma)$. The causal pair $(X\!\to\!Y,\,Y\!\not\to\!X)$ is hard for linear and Granger-style methods, which see strong autocorrelation in both series. Indeed, for moderate levels of noise $\sigma^2=0.4$, these methods fail to detect the correct causal links \cite{CampsValls23physcausaldiscovery}. Here, we investigate whether our Bayesian symbolic regression approach can detect the correct causal relationships and governing equations at different noise levels (Fig.~\ref{fig:shape}). The levels of noise we investigate go from low, where autocorrelation and cross-correlation between both time series are strong (Fig.~\ref{fig:shape}a), to very high, where autocorrelation in the $x_t$ time series is much weaker, and the dynamics of $y_t$ is dominated by some sporadic large bursts (Fig.~\ref{fig:shape}d). In all cases, we let causalBMS explore time lags $\tau=1, 2, 3$ to evaluate whether autocorrelation in the time series leads to failure to distinguish direct from indirect temporal effects or other forms of lag misidentification.

As we show in Fig.~\ref{fig:shape}e-h, causalBMS correctly and unequivocally identifies causal parents and lags at all levels of noise, with lagged causal scores ${\pi}_{k\to i}(\tau)$ close to one when a causal link $k\to i$ exists at $\tau$, and close to zero when it does not. Only when the autocorrelation in the $x_t$ series becomes very weak, does the corresponding score decrease to ${\pi}_{X\to X}(1)=0.74$, still clearly above 0.5 and above the true negative edges.

Our causalBMS draws from the posteriors over equations for $x_{t}$ and $y_{t}$ separately and reads directionality from the inclusion frequencies in Eq.~\eqref{eq:inclusion_marg}. The ensemble assigns lag $\tau\!=\!1$ in $X$ a high inclusion probability inside the equation for $y_{t}$, and $Y$ a very low inclusion probability inside the equation for $x_{t}$, recovering the correct causal effect. Beyond directionality, the method recovers, or at least approximates, the functional form of the interaction (Fig.~\ref{fig:shape}i-l). At low noise ($\sigma=0.04$), the true $f_i^*$ is well approximated within the observed range of $x(t)$ (Fig.~\ref{fig:shape}i), which is relatively narrow and prevents causalBMS from learning the limiting behavior and, thus, the exact functional form $f_i^*$. Conversely, at high noise ($\sigma \ge 4$, Fig.~\ref{fig:shape}k,l) the signal-to-noise ratio is so low that $\sigma$ is above $\sigma^{\times}$, as defined in Eq.~\eqref{eq:phase_bound}, and the true $f_i^*$ cannot be learned by any method \cite{fajardo2023fundamental}. Despite this, the method still detects the correct causal effect. Finally, moderate noise levels ($\sigma=0.4$) lead to exact recovery of the $f_i^*$ (Fig.~\ref{fig:shape}j) because, in this regime, the signal-to-noise ratio is high enough to bring the data into the learnable regime and, at the same time, the observed range of $x(t)$ is large enough that the limiting behavior of $f_i^*$ is visible.

Thus, this example illustrates three important and somewhat counterintuitive features of causal discovery based on Bayesian symbolic regression: (i) causalBMS has no problem in detecting unidirectional causal relationships, even in scenarios where the state of the art in causal discovery methods fail \cite{CampsValls23physcausaldiscovery}; (ii) causal effects are detectable even in the unlearnable symbolic regression phase; and (iii) more noise can actually increase the signal-to-noise ratio by increasing the range of observed values in the time series.

\subsection{CauseMe benchmarks}
\label{sec:causeme}

\begin{figure}[t!]
\centering
\includegraphics[width=\columnwidth]{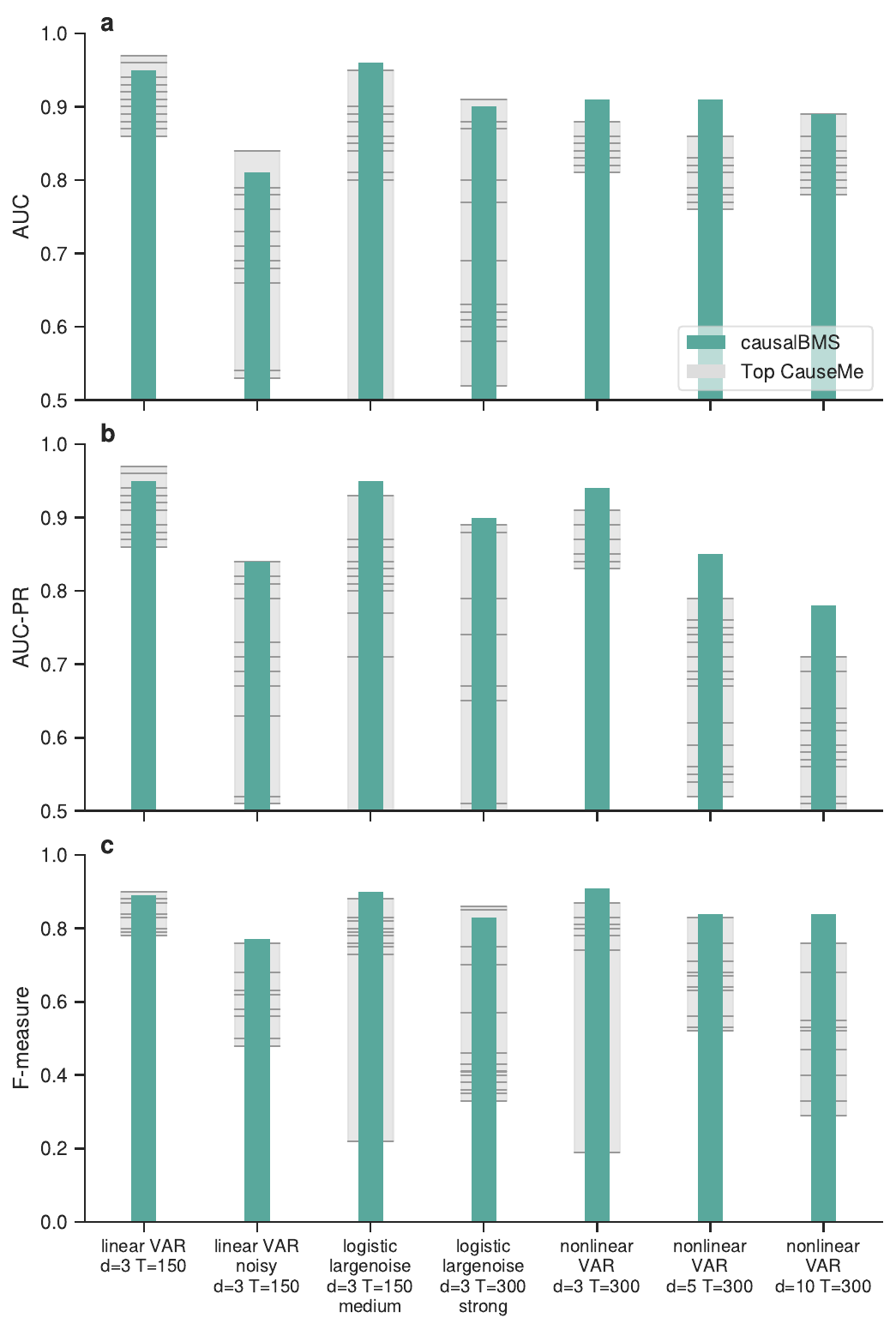}
\caption{\textbf{Performance of causalBMS on the CauseMe benchmark.} For the experiments described in \S\ref{sec:causeme}, we show: (a) area under the ROC curve; (b) area under the precision--recall curve; and (c) F1-score. The green bars are the causalBMS scores; the grey ranges span the top-20 entries reported by CauseMe for each dataset. causalBMS sits at or near the top of every column and consistently falls within the upper range of competing methods, especially for the nonlinear and strong-noise variants.}
\label{fig:causeme}
\end{figure}

Next, we evaluate causalBMS on the CauseMe benchmark suite curated for the NeurIPS 2019 Causality~for Climate competition~\cite{runge2020causality,runge2019detecting}. The suite stresses three orthogonal axes: dynamical complexity (non-linearity of the dynamics and dimensionality $d$ of the system), sample size $T$, and signal-to-noise ratio. Here, we consider the following datasets:
\begin{itemize}
\item {\bf Linear baselines.} The \texttt{linear-VAR} family ($d\!=\!3$, $T\!=\!150$) provides a controlled baseline of linear models; the \texttt{linear-VAR\_noisy} configuration adds strong exogenous noise for the relatively small samples available.

\item {\bf Nonlinear and chaotic systems.} Coupled logistic maps (\texttt{logistic-largenoise}, $d\!=\!3$, medium and strong noise) test the high-noise transitional regime, while \texttt{nonlinear-VAR} with polynomial dependencies probes scalability across system dimensionality $d\!\in\!\{3,5,10\}$ at $T\!=\!300$.
\end{itemize}
Each dataset consists of 250 time-series sets from which the causal graph must be inferred. We compare causalBMS with the 20 top performers for each dataset in the CauseMe leaderboard (Fig.~\ref{fig:causeme}).

Across all setups, causalBMS is consistently among the top performers in F1-score, area under the precision--recall curve (AUC-PR), and area under the ROC curve (AUC) (Fig.~\ref{fig:causeme}). Indeed, causalBMS is the top performer in 13 out of 21 dataset-metric pairs, the second-best in another four pairs, and consistently close to the top performer when it is not the best. Our approach works particularly well in nonlinear systems. The robustness in the strong-noise variants also supports using the score based on the probabilistic approach and the complete posterior distribution.

% =======================================================
\subsection{A biogeoscience case study: temperature, productivity, and respiration}
\label{sec:reco}

\begin{figure*}[t!]
\centering
\includegraphics[width=\textwidth]{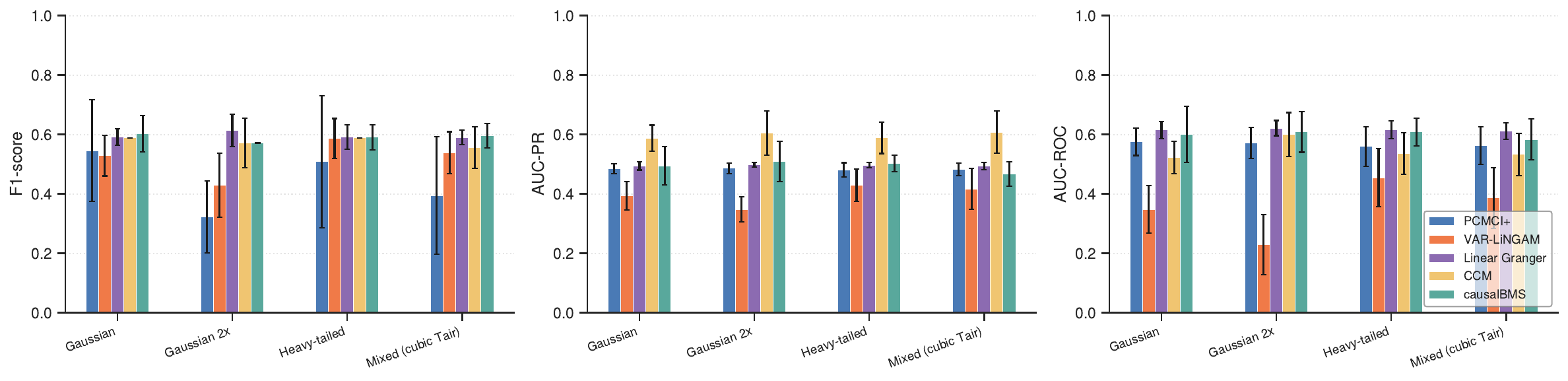}
\caption{\textbf{Multi-method comparison on the biogeoscience SCM of Eq.~\eqref{eq:reco_scm}.} F1-score (left), AUC-PR (middle), and AUC-ROC (right) on the cross-influence entries of the lag-one summary graph, averaged over twenty replicates per noise regime; error bars are one standard deviation. Regimes (left to right inside each panel): Gaussian fluctuations at the reference scale, Gaussian fluctuations at twice the reference scale, Student-$t$ fluctuations with $\nu\!=\!4$ degrees of freedom and a mixed regime with an additional cubic term on the $T_{\mathrm{air}}$ fluctuations. causalBMS (green) is among the leading methods for F1-score and AUC-ROC and often ties for the best performer on these metrics. CCM (yellow) is the strongest method for AUC-PR throughout. In AUC-ROC, a linear Granger-causality test extended with contemporaneous terms (purple) matches or slightly exceeds every other method, including causalBMS, in all four regimes. VAR-LiNGAM (red) is consistently the weakest method in AUC-ROC and AUC-PR, while PCMCI$^{+}$ (blue) shows more erratic performance across noise regimes in F1-score.}
\label{fig:reco_panel}
\end{figure*}
Finally, we consider a four-variable causal graph rooted in ecosystem physiology, which we choose because it violates some assumptions of causalBMS and is therefore particularly challenging for our approach. The structural model follows the simulated case study of Runge et al.~\cite{runge2023causal}, in which incoming shortwave radiation $R_{g}$ drives gross primary productivity $\mathrm{GPP}$ through air temperature $T_{\mathrm{air}}$, and ecosystem respiration $R_{\mathrm{eco}}$ responds nonlinearly to $T_{\mathrm{air}}$ and $\mathrm{GPP}$ through a $Q_{10}$-like kernel. The system encodes the standard light-temperature partitioning used to separate net ecosystem exchange into assimilation and respiration~\cite{reichstein2005separation,MAHECHAetal2010c} and is a controlled surrogate for the eddy-covariance setting routinely analyzed on FLUXNET sites~\cite{pastorello2020fluxnet2015}. The lag-one summary graph contains eight directed edges, including four cross-influences ($R_{g}\!\to\!T_{\mathrm{air}}$, $R_{g}\!\to\!\mathrm{GPP}$, $T_{\mathrm{air}}\!\to\!\mathrm{GPP}$, $T_{\mathrm{air}}\!\to\!R_{\mathrm{eco}}$, $\mathrm{GPP}\!\to\!R_{\mathrm{eco}}$) and four autoregressive self-edges.
The discrete-time SCM is, with seasonal phase $\phi_{t}=t\pi/365$ and i.i.d.\ fluctuations $\eta^{i}_{t}$,
\begin{align}
R_{g}(t) &= 280\sin^{2}\phi_{t} + 50\,|\!\sin\phi_{t}|\,\eta^{R_{g}}_{t}, \label{eq:reco_scm}\\
T_{\mathrm{air}}(t) &= 0.8\,T_{\mathrm{air}}(t\!-\!1) + 0.02\,R_{g}(t) + 5\eta^{T}_{t},\nonumber\\
\mathrm{GPP}(t) &= \left| 0.2\,\mathrm{GPP}(t\!-\!1) + 0.002\,T_{\mathrm{air}}(t) R_{g}(t) + 3\eta^{G}_{t} \right| ,\nonumber\\
R_{\mathrm{eco}}(t) & = \left| 0.3\,R_{\mathrm{eco}}(t\!-\!1) + 0.9\,\mathrm{GPP}(t)\,0.8^{0.12(T_{\mathrm{air}}(t)-15)} + 2 \eta^{R}_{t} \right|. \nonumber
\end{align}

Each simulation produces $T\!=\!800$ daily samples followed by a low-radiation seasonal mask that retains only the active season. 
We compare five methods spanning the constraint-, structural-, dynamics- and regression-based families: PCMCI$^{+}$ with partial-correlation independence tests at $\tau_{\max}=1$~\cite{runge2019detecting,krich2020estimating}, VAR-LiNGAM~\cite{shimizu2006linear,hyvarinen2010estimation}, convergent cross mapping (CCM)~\cite{Sugihara12}, a linear Granger-causality test extended with contemporaneous (lag-zero) terms to allow for instantaneous influences in Eq.~\eqref{eq:reco_scm}~\cite{Shojaie2021-ge}, and causalBMS.

We evaluate on four noise regimes: Gaussian fluctuations at the reference scale $\sigma\!=\!1$ and at $\sigma\!=\!2$, heavy-tailed Student-$t$ fluctuations with $\nu\!=\!4$ degrees of freedom, and a mixed regime in which the temperature fluctuation includes an additive cubic distortion $0.25\varepsilon^{3}$ as in the Runge example~\cite{runge2023causal}, modeling occasional extremes. Each method is run on five independent replicates per regime; we report the F1-score, the area under the precision-recall curve (AUC-PR) and the area under the ROC curve (AUC-ROC) computed on the twelve off-diagonal entries of the lag-one summary graph (Fig.~\ref{fig:reco_panel}).

We note that this experiment is singularly challenging for the causalBMS for at least two reasons. First, in some experiments, the additive noise terms are not Gaussian. Second, and more significantly, functional identification is misspecified in Eq.~\eqref{eq:reco_scm} because of the contemporaneous (lag zero) influence terms. For example, temperature $T_{\rm air}$ is causally affected only by radiation $R_g$, but because of the lack of temporal resolution, the effect of $R_g$ is measured as instantaneous. Similarly, the causal effect of $T_{\rm air}$ and $R_g$ on ${\rm GPP}$ and $R_{\rm eco}$ is also measured as instantaneous. Therefore, $T_{\rm air}$ can be written as a function of $R_g$, but also as a function of ${\rm GPP}$ and of $R_{\rm eco}$.

Despite these intrinsic difficulties, causalBMS remains competitive with the strongest methods on most of the metrics in Fig.~\ref{fig:reco_panel}. In F1-score, causalBMS sits within the leading cluster together with CCM and linear Granger across all regimes and is often the best or close-to-best performer. In AUC-PR, causalBMS is in the second tier, alongside PCMCI$^{+}$ and linear Granger, all below CCM, which is the strongest method on this metric in all regimes. On AUC-ROC, linear Granger with contemporaneous terms matches or exceeds every method on average in all four regimes, but causalBMS is always a very close second and statistically indistinguishable from the top performer. VAR-LiNGAM is consistently the weakest method in AUC-ROC across all four regimes, reflecting its inability to represent the multiplicative and exponential terms. In F1-score, however, its performance is comparable to, and in some regimes higher than, PCMCI$^{+}$, whose linear partial-correlation test is itself much more uncertain across noise regimes than either causalBMS or linear Granger.

That a simple contemporaneous-term linear Granger test matches causalBMS in edge detection reflects a property of this benchmark: the mechanisms themselves in Eq.~\eqref{eq:reco_scm} are close enough to linear that a correctly-specified, low-parameter significance test is a statistically efficient detector of an edge's presence.

\begin{figure*}[t!]
\centering
\includegraphics[width=\textwidth]{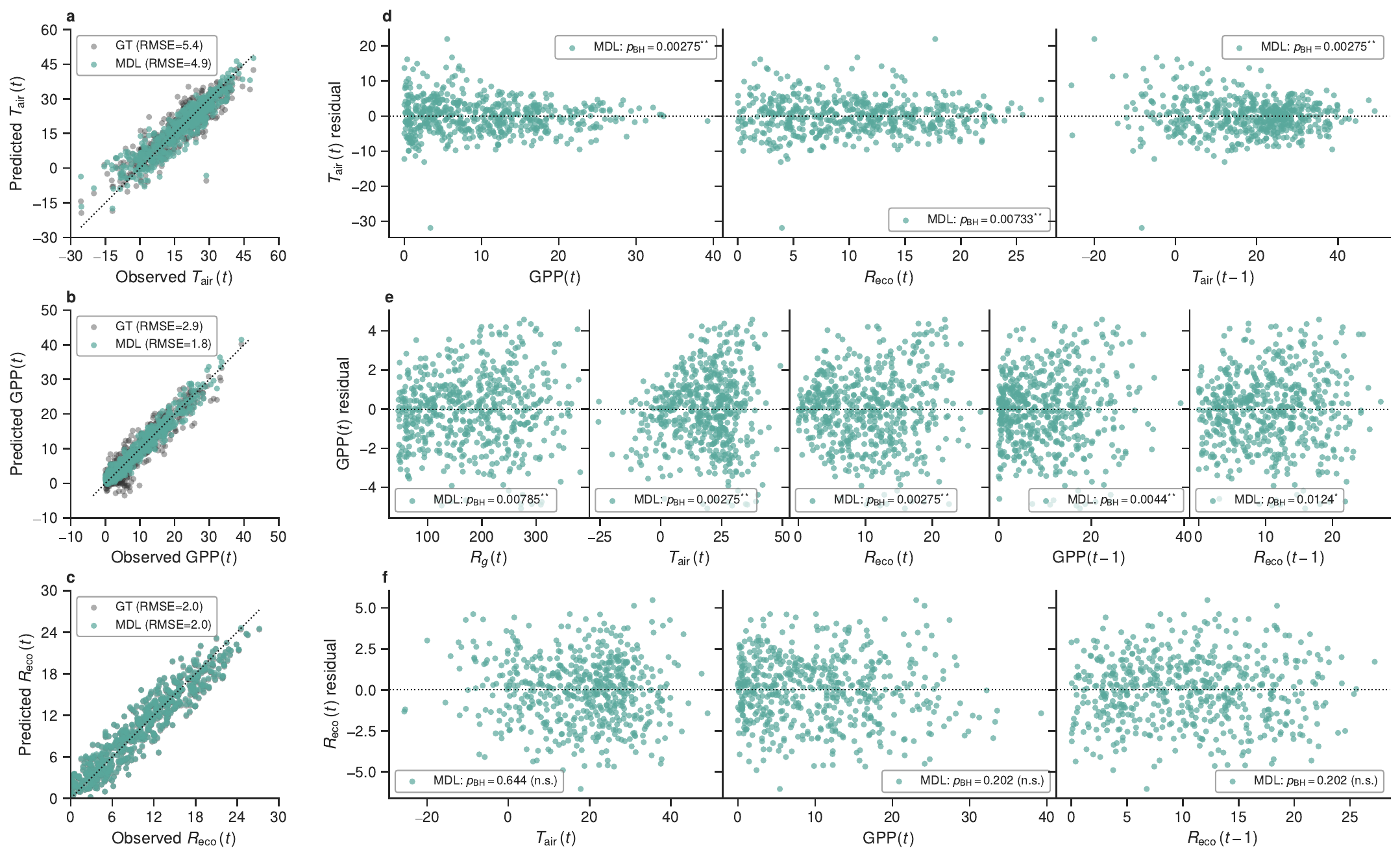}
\caption{\textbf{Properties of minimum description length models identified by the causalBMS for the biogeoscience model of Eq.~\eqref{eq:reco_scm}.} After obtaining the MDL model for each target with the causalBMS, we generate test data with the same SCM model of Eq.~\eqref{eq:reco_scm} and test the predictive abilities and correlation structure of the models. (a-c) Predicted versus observed values of the target variables for the MDL model (green) and the ground-truth model (grey). (d-f) Correlations between residuals and predictors for each target variable and each predictor identified by the MDL model. We report the BH-corrected $p$-values for a permutation test with HSIC assessing remaining correlation between residuals and chosen predictors.}
\label{fig:reco_panel_2}
\end{figure*}

\begin{table*}[t!]
\caption{\textbf{Equations recovered by causalBMS on the biogeoscience SCM.} For each target, we show the ground truth model and the MDL expression identified by the causalBMS in the reference Gaussian-noise regime.}
\label{tab:reco_eqs}
\centering
\small
\setlength{\tabcolsep}{1pt}
\begin{tabular}{@{}l p{5.8cm} p{11.5cm}}
\toprule
 & True mechanism (from Eq.~\eqref{eq:reco_scm}) & MDL causalBMS expression\\
\midrule
$T_{\mathrm{air}}$ \hspace{1mm} & $0.8\,T_{\mathrm{air}} (t\!-\!1)$ $+\,0.02\,R_{g}(t)$ & $0.82 \, T_{\rm air}(t-1) + 0.84 \, {\rm GPP}(t) - 0.023 \, T_{\rm air}(t-1) \, R_{\rm eco}(t)$\\[2pt]

$\mathrm{GPP}$ \hspace{1mm} & $0.2\,\mathrm{GPP}(t\!-\!1)$ $+\,0.002\,T_{\mathrm{air}}(t)\,R_{g}(t)$ & $1.1 \cdot 10^{-5} \left(620 \, R_{\rm eco}(t) + R_g(t) T_{\rm air}(t)\right) \left({\rm GPP}(t-1) + R_{\rm eco}(t) - 2.0 \, R_{\rm eco} (t-1) + T_{\rm air}(t) + 64 \right)$\\[2pt]

$R_{\mathrm{eco}}$ \hspace{1mm} & $0.3\,R_{\mathrm{eco}}(t\!-\!1) + 0.9\,\mathrm{GPP}(t)\,0.8^{0.12(T_{\mathrm{air}}(t)-15)}$ & $0.32 \, R_{\rm eco} (t-1) + 1.3 \, {\rm GPP}(t) \, {0.97}^{T_{\rm air}(t)} $ \\[2pt]
\bottomrule
\end{tabular}
\end{table*}
Unlike PCMCI$^{+}$, VAR-LiNGAM, linear Granger and CCM, which return only edge weights, causalBMS returns the closed-form expressions that produced those weights, together with a diagnostic (Fig.~\ref{fig:reco_panel_2}d-f) that identifies, without reference to the ground truth, which of its recovered mechanisms remain incompletely specified. Table~\ref{tab:reco_eqs} compares the MDL expression obtained by the causalBMS for each target to the ground-truth mechanism in Eq.~\eqref{eq:reco_scm}. 

For $R_{\mathrm{eco}}$, the MDL expression not only captures the causal parents correctly but also matches exactly the ground truth expression, up to some error in the coefficients. This optimal outcome is possible because $R_{\rm eco}$ has the only functional dependency that is not misspecified in Eq.~\eqref{eq:reco_scm}.
For $T_{\rm air}$ and ${\rm GPP}$, the MDL models also capture the leading terms in the ground-truth causal mechanism---$T_{\rm air}(t-1)$ and the product $R_g(t) \, T_{\rm air}(t)$, respectively. However, the models also include other terms with incorrect causal parents. The fact that the causalBMS prefers models that are more complex than the ground truth leads us to hypothesize that they must be more predictive than the ground truth. Indeed, as per Eq.~\eqref{eq:bic}, the only way a more complex model (lower $\log p(f_i)$) is acceptable is if it reduces the BIC term of the description length even more.
In Fig.~\ref{fig:reco_panel_2}a-c, we show that this is indeed the case---the identified MDL expressions have, for all three targets, lower prediction error than the ground-truth expressions. This counterintuitive result is a consequence of the misspecification in Eq.~\eqref{eq:reco_scm}. Consider, for example, $T_{\rm air}$. Since, functionally, $T_{\rm air}$ can be solved for analytically from ${\rm GPP}$ and $R_{\rm eco}$, it is in general better to use a combination of these expressions to estimate $T_{\rm air}$, since such a combination will tend to cancel the noise terms and therefore lead to more accurate estimates.

We now test whether each recovered mechanism fully accounts for its own selected inputs. For this, we compute the Hilbert-Schmidt independence criterion (HSIC) between the residuals of each target and every predictor chosen in the respective MDL expression (Table~\ref{tab:reco_eqs}). We assess the significance using a permutation test. We run 1000 permutations and apply the Benjamini-Hochberg (BH) correction to account for multiple testing. 
This assessment is a diagnostic for the adequacy of the functional form that could apply in principle to the fitted equation of any method, regardless of how its inputs were selected. For $T_{\rm air}$ and $\mathrm{GPP}$, every predictor entering their MDL expressions remains significantly dependent on the residual after correction ($p_{\rm BH} \le 0.0124$, Fig.~\ref{fig:reco_panel_2}d-e). This shows that neither expression fully captures the target's dependence on its own selected inputs. This is consistent with their larger deviation from the ground-truth mechanism in Table~\ref{tab:reco_eqs}. At the same time, none of the three predictors of the $R_{\rm eco}$ expression show significant residual dependence ($p_{\rm BH} \ge 0.202$, Fig.~\ref{fig:reco_panel_2}f). This agrees with its near-exact recovery of both the ground-truth mechanism and RMSE (Fig.~\ref{fig:reco_panel_2}c). Because this diagnostic requires no access to the ground-truth model, it offers a practical, general-purpose check on whether a fitted expression (from causalBMS or any regression-based causal discovery method) is fully specified, complementing aggregate error metrics such as RMSE with a finer-grained structural test of residual adequacy.

\section{Discussion}
\label{sec:discussion}
We show that casting causal discovery as Bayesian symbolic regression turns the BMS posterior over expressions into a causal score via variable-inclusion frequencies. The score preserves the description length semantics of the BMS, extends naturally to many candidate parents and lagged predictors, and remains informative through the learnability transition characterized in Ref.~\cite{fajardo2023fundamental}. Because each sampled expression directly enumerates its predictors, the score does not require comparing candidate factorizations or selecting a single best equation, and exploits the full posterior of plausible mechanisms.

Three limitations qualify the present results. 
First, identifiability is conditional on closed-form representability of the true mechanism; outside this class the inclusion frequencies remain well defined as a feature-selection signal but lose the consistency guarantee of Proposition~\ref{prop:consistency}. 
Second, the BMS sampler scales with the number of candidate parents; current implementations comfortably handle the benchmark dimensions ($d\!\le\!10$, $T\!\le\!300$) but require additional ingredients to reach the high-dimensional settings of some real data. 
Third, the noise threshold in Eq.~\eqref{eq:phase_bound} depends on quantities ($\langle\delta^{2}\rangle$, $\Delta^{*}_{\!M}$) that involve the unknown true model and must be estimated from data; we use it as a diagnostic rather than as a hard guarantee.

Several extensions are natural and are likely to alleviate these limitations. Combining causalBMS with constraint-based skeleton discovery~\cite{Spirtes2000,runge2019detecting} would exploit the strengths of both families: independence tests prune the candidate parent sets, and symbolic regression orients edges and recovers mechanisms. A similar approach is exploited by the recent TOPIC framework~\cite{xu2025information}, which infers a topological order from algorithmic mutual information and identifies edges that compress the data under a candidate ordering; embedding causalBMS as the per-node scoring `oracle' in such a topological pipeline would extend the method to higher dimensions while preserving its mechanism-level output. Embedding causalBMS inside hybrid modeling pipelines would provide explicit mechanism candidates for downstream physical evaluation, and the ODE-SCM correspondence of Ref.~\cite{mooij2013ordinary} suggests a route from discovered difference equations to interventional predictions. \\

In any case, we have shown that causal BMS often performs better than the state of the art. Moreover, the principal scientific gain of causalBMS over the alternatives is that it not only provides edge ranking in the causal graph, but also the explicit functional form of the surviving expressions, which is a much richer description of mechanisms, allows to test for remaining correlations, and is typically an additional object of interest in science.

% --------------------------------------------------------------
\acknowledgements

This work is supported by the European Research Council (ERC) through grant USMILE (grant agreement No.~855187); the European Commission through the Horizon Europe project ELIAS (grant agreement No.~101120237); projects PID2022-142600NB-I00, PID2022-142600NB-I00, and PID2025-176129NB-I00 from MCIN / AEI / 10.13039 / 501100011033; and by the Government of Catalonia (2021SGR-633).

% --------------------------------------------------------------
\section*{Author contributions}

RG, KHC, MSP, and GCV designed the research. RG, KHC, SP, and GCV wrote code and run simulations. All authors discussed and interpreted the results, worked on the visualizations, contributed to writing and revising the manuscript, and approved the final version.

% --------------------------------------------------------------
\section*{Code and data availability}

The code for the BMS is openly available \cite{bmscode}. Data for the CauseMe benchmark are available from the \href{https://causeme.uv.es}{CauseMe platform} \cite{runge2020causeme}. All other data used in the manuscript were synthetically generated as described in the text.

% --------------------------------------------------------------
\section*{Competing interests declaration}

The authors declare that they have no competing interests.

% BIBLIOGRAPHY -----------
%\bibliographystyle{apsrev4-2}
\bibliography{references}

\end{document}